\documentclass{article}
\pdfoutput=1
\usepackage{spconf}
\usepackage{amsmath}
\usepackage{graphicx}
\usepackage{amssymb}
\usepackage{multirow}
\usepackage{array}
\usepackage{indentfirst}
\usepackage{svg}
\usepackage{float}
\usepackage[colorlinks,citecolor=green,urlcolor=blue,bookmarks=false,hypertexnames=true]{hyperref} 

\usepackage{authblk}

\title{LP-IOANet: efficient high resolution document shadow removal}

\begin{document}
\name{Konstantinos Georgiadis\textsuperscript{1}*, M. Kerim Yucel\textsuperscript{2}*, Evangelos Skartados\textsuperscript{1}*, Valia Dimaridou\textsuperscript{1}, \\
Anastasios Drosou\textsuperscript{1}, Albert Sa\`a-Garriga\textsuperscript{2},
Bruno Manganelli\textsuperscript{2}\thanks{* The first three authors contributed equally.}}
\address{\textsuperscript{1} CERTH, Information Technologies Institute, Thessaloniki, Greece  \\
\textsuperscript{2} Samsung Research UK}
\maketitle

\begin{abstract}
Document shadow removal is an integral task in document enhancement pipelines, as it improves visibility, readability and thus the overall quality. Assuming that the majority of practical document shadow removal scenarios require real-time, accurate models that can produce high-resolution outputs in-the-wild, we propose \textbf{L}aplacian \textbf{P}yramid with \textbf{I}nput/\textbf{O}utput \textbf{A}ttention \textbf{Net}work (\textbf{LP-IOANet}), a novel pipeline with a lightweight architecture and an upsampling module. Furthermore, we propose three new datasets which cover a wide range of lighting conditions, images, shadow shapes and viewpoints. \textcolor{black}{Our results show that we outperform the state-of-the-art by a 35\% relative improvement in mean average error (MAE), while running real-time in four times the resolution (of the state-of-the-art method) on a mobile device.}

\end{abstract}
\begin{keywords}
Shadow Removal, Document Enhancement, Super-Resolution
\end{keywords}
\vspace{-4mm}
\section{Introduction} \label{introduction}
\vspace{-1mm}

\noindent  The wide spread use of mobile phone cameras has made document digitization significantly practical. Using mobile phone cameras often leads to issues like distortion, blur, noise and shadows cast on documents. Document shadow removal task aims to remove shadows cast on document images in a visually pleasant manner. Despite the recent advances  \cite{bako2016removing,jung2018water,kligler2018document,lin2020bedsr}, there are issues with existing methods. First, the majority of the existing methods do not often aim for a lightweight solution. Second, most methods do not operate at high resolutions. Third, document images come in many forms, such as text-only colorless documents and colored/figure-heavy documents, which necessitates good performance in-the-wild. A recent work \cite{lin2020bedsr} addresses document shadow removal in an end-to-end manner, however, it does so with a large model that does not operate at high resolutions.

\begin{figure*}[!ht]
  \centering
    \includegraphics[width=\textwidth]{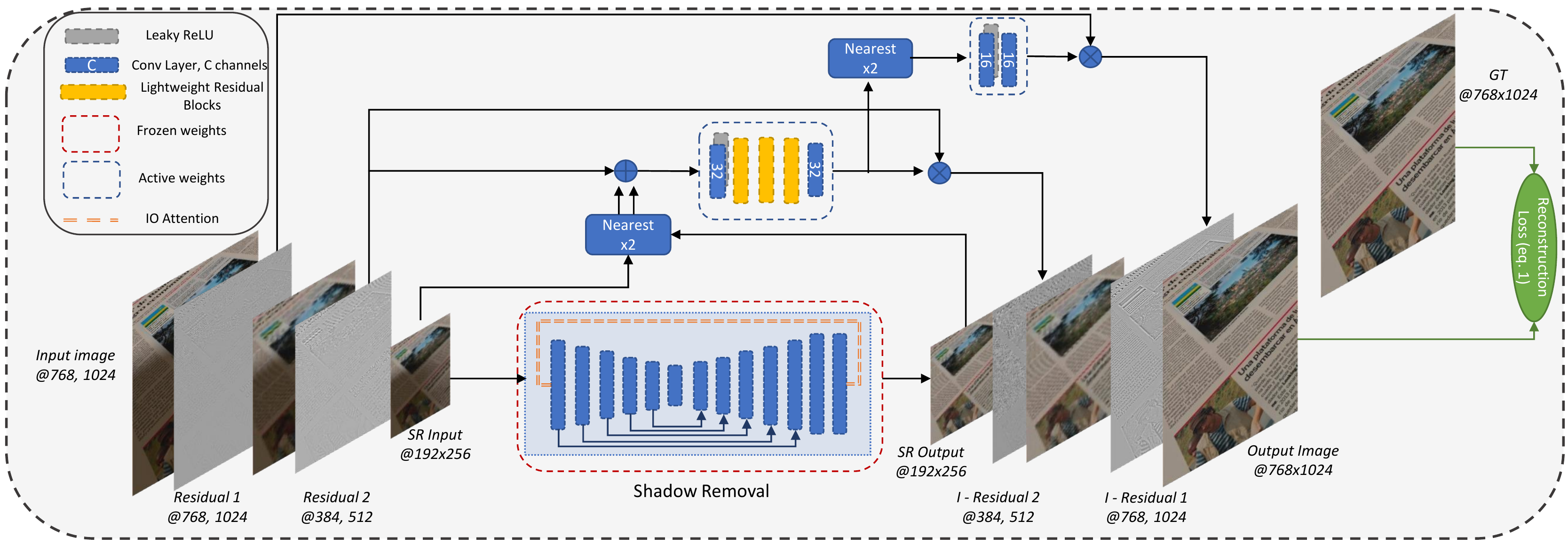}
  \vspace{-8mm}
  \caption{Our LP-IOANet pipeline. Following the training of our shadow removal network (red dashed lines) in low-resolution (see Figure \ref{fig:networks}), we freeze it and train our lightweight upsampling module (dashed blue lines) on our proposed A-BSDD dataset.}
  \label{fig:overall_diagram}
  \vspace{-4mm}
\end{figure*}

The contributions of our work are as follows: we propose i) IOANet, a document shadow removal network with input/output attention for real-time operation, ii) a lightweight upsampling module that encapsulates IOANet, letting us operate at high resolutions, iii) three new datasets which cover various lighting conditions, document types and viewpoints and iv) a two-stage training pipeline that lets us leverage any low-resolution dataset for improved generalization. Our LP-IOANet comfortably outperforms the state-of-the-art, runs in real-time on a mobile device in 4 times the resolution of the state-of-the-art. The state-of-the-art method runs out of memory, thus can not be run, even on a 24GB VRAM desktop GPU. The diagram of LP-IOANet is shown in Figure \ref{fig:overall_diagram}.

\vspace{-3mm}

\section{Related Work} \label{related_work}
\vspace{-1mm}

\noindent \textbf{Document Shadow Removal and Upsampling} Earlier works based on intrinsic images \cite{yang2012shadow,brown2006geometric} and hand-crafted methods \cite{jung2018water,bako2016removing,kligler2018document} are built on simplifying assumptions and fail to perform effectively. Most recently, a deep-learning based solution BEDSR is proposed \cite{lin2020bedsr}. BEDSR assumes document images have a dominant background color, and explicitly estimates this via a background estimation module. This is used to create an attention map of background/foreground pixels, which acts as a shadow mask. Input image and the attention map are fed to an encoder/decoder to remove the shadows. On the other hand, several document specific super-resolution methods, based on CNNs \cite{pandey2017language} or GANs \cite{peng2020building}, are shown to be effective and also improve OCR performance. 

\noindent \textbf{Document Shadow Removal Datasets.} Existing real-life document shadow removal datasets are quite small  \cite{bako2016removing,jung2018water,kligler2018document,lin2020bedsr} and thus not suitable for training purposes. Synthetically casting shadows on shadow free images is arguably more practical than creating large-scale real-life datasets, as synthetic datasets can eliminate intrinsic data errors (on non-shadow regions), simulate various lighting/occluder settings and can automate labelling process entirely. SDSRD is the largest synthetic dataset formed of 8.3K triplets, created from 970 unique images \cite{lin2020bedsr}, however, it is not publicly available and even larger datasets are still desirable.

In comparison to the current state-of-the-art BEDSR \cite{lin2020bedsr}, our work has several key advantages: we i) do not need the background color labels during training, ii) address high resolution output requirement explicitly, iii) enlarge data distribution via our new datasets and iv) do not use large, complex architectures. Our pipeline is also a prime candidate for optimizations via pruning/quantization as it uses simple building blocks, whereas BEDSR requires gradient information \cite{selvaraju2017grad} in inference, which can be hard to obtain on resource-constrained environments like mobile phones. Such optimizations are interesting for future work, but not within our scope.

\vspace{-2mm}

\section{Methodology}
\label{methodology}

\vspace{-2mm}
\subsection{Shadow Removal}
\begin{figure*}[htbp]
  \centering
      \includegraphics[width=\textwidth]{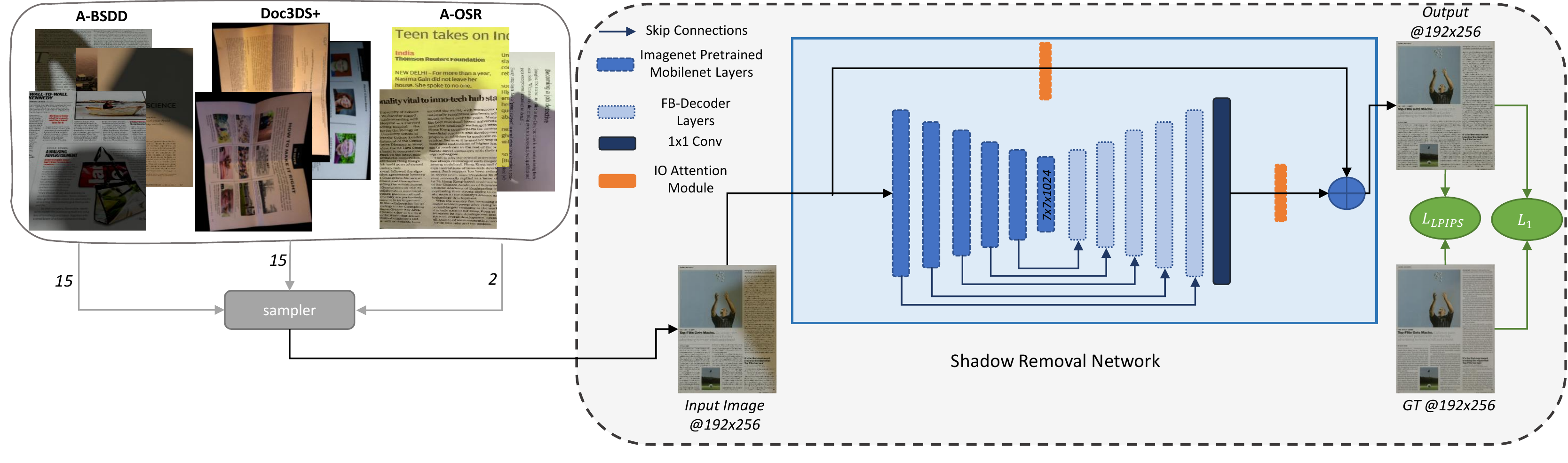}
  \vspace{-8mm}
  \caption{Our IOANet shadow removal network. Using all three datasets we propose, we train our network on low-resolution images using a combination of L1 and LPIPS losses. This training stage is followed by stage 2 (see Figure \ref{fig:overall_diagram}).}
  \label{fig:networks}
  \vspace{-4mm}
\end{figure*}

\noindent  
Our aim is to mimic the high-fidelity results of two-network setups (removal and refinement) \cite{le2020physicsbased}, but by using a single, efficient network that implicitly localizes and removes shadows. We base our architecture on \cite{yucel2021real} due to its strong accuracy/runtime performance and propose to use lightweight attention modules \cite{hou2021coordinate,yucel2023lra} over the input and output (IOA) of the network, and sum their results via a long residual connection from input to output. IOA has the advantage of being parallelizable (i.e. input attention is executed concurrently with the  network), makes the network focus on the shadow areas only (i.e. non-shadow areas are copied by the long residual connection) and introduces additional capacity for blending/color-correction with minimal computational overhead. We call the resulting architecture IOANet; its architecture and training details are shown in Figure \ref{fig:networks}.

\vspace{-2mm}
\subsection{Efficient Upsampling}

\noindent \textbf{Preliminaries.} The visual quality of the output is important in documents, since high frequency components (i.e. text) must be preserved after generation and upsampling. We set the target resolution for the shadow removal network to (192$\times$256), which is reminiscent of the aspect ratio of documents. Instead of naively running the network at high-resolution, we aim to efficiently upsample the shadow-removed image four times to high-resolution (768$\times$1024).

\noindent \textbf{The proposed method.} Laplacian Pyramid Networks \cite{liang2021high} decompose an image into a Laplacian pyramid \cite{burt1987laplacian}, where low-frequency components are fed to an image-to-image translation network in low resolution. High frequency components are adaptively refined via a mask learning network based on all frequency components. This mask is upsampled and finetuned for each resolution level, and all components are used to reconstruct the high resolution output.

We use a 2-level pyramid where IOANet operates on low resolution (192$\times$256) images. Unlike the original work \cite{liang2021high}, we train IOANet with low-resolution images and then the rest of the network with high-resolution images (see Section \ref{sec:imp_details}). The residual refinement network operates on the intermediate resolution of (384$\times$512).  This network, in its original version, leads our pipeline to have 22.8 GFLOPs complexity. Unlike \cite{liang2021high}, we implement the residual refinement network with cheap, depthwise separable convolutions, resulting in 3.82 GFLOPs (called LPTN-lite). We further decrease the width of the network and achieve 1.47 GFLOPs; this is our LP-IOANet pipeline. LP-IOANet is shown in Figure \ref{fig:overall_diagram}.

\vspace{-4mm}
\subsection{Datasets}

\noindent The largest dataset in the literature (SDSRD \cite{lin2020bedsr}) is not publicly available, therefore we opt to create our own synthetic datasets. In addition to being able to train on a more diverse dataset, our new datasets allow us to evaluate on a larger distribution, giving us a better insight on models' performance.

\noindent \textbf{BSDD.} We follow the principles of \cite{lin2020bedsr} and create Blender Synthetic Dataset (BSDD) using 1328 unique images. BSDD has 3863 high resolution triplets, split into 3477 and 386 images for training and testing, respectively.

\noindent \textbf{Doc3DS+.} We leverage the Doc3DShade dataset \cite{DasDocIIW20}, which is formed of 90K image triplets of shadow images with colored backgrounds, white balanced shadow images and albedo document images. Training a model using colored shadowed images as input and albedo as output may result in neglecting the paper color, whereas using white balanced shadowed as input may restrict the model's view to white paper only. We extract the background color from the input image using color clustering and reapply it on the albedo document images to alleviate such issues. We rotate the resulting images to be closer to our desired A4 document resolution. 

\noindent \textbf{Augmenting the datasets.} A shadow area is not just a darkened version of the original image; natural shadows tend to have different colors and illuminations. Following \cite{le2020physicsbased, Le_2019_ICCV}, we apply illumination augmentation. Furthermore, we also modify the colour values of the shadows, which gives us a more diverse distribution. With this procedure, we augment the train split of BSDD and the entire OSR \cite{osrDataset} dataset and use the augmented versions \textit{A-BSDD} and \textit{A-OSR} in our experiments. Details of our datasets are shown in Table \ref{tab:datasetsOverview}.

\begin{table}[]
\tiny
\resizebox{0.485\textwidth}{!}{%
\begin{tabular}{l|c|c|c}
Dataset & Size & Unique Images & Resolution \\ \hline
A-BSDD & 24082 & 1328 & High \\
Doc3DS+ & 71595 & 9393 & Low \\
A-OSR & 1410 & 23 & Low
\end{tabular}%
}
\vspace{-4mm}
\caption{Overview of the datasets.}
\label{tab:datasetsOverview}
\vspace{-5mm}
\end{table}

\begin{table*}[htbp!]
\resizebox{\textwidth}{!}{%
\begin{tabular}{l|ccc|c|c|c}
\multirow{2}{*}{Method} & \multicolumn{3}{c|}{BSDD} & Runtime & Memory & \multirow{2}{*}{GFLOPs} \\ \cline{2-6}
 & \multicolumn{1}{c|}{MAE $\downarrow$ } & \multicolumn{1}{c|}{PSNR $\uparrow$} & SSIM $\uparrow$ & ms  & GB  &  \\ \hline
Input Images - No Removal & \multicolumn{1}{c|}{7.2256 / 2.0526 / 24.156} & \multicolumn{1}{c|}{23.98 / 13.63 / 13.56} & 0.95 &  - & - & - \\ \hline
BEDSR \cite{lin2020bedsr} & \multicolumn{1}{c|}{2.8321 / 2.1534 / 5.0535} & \multicolumn{1}{c|}{34.25 / 13.58 / 13.94} & 0.98 &  119.0 & 2.30 & 550 \\

IOANet w/o attention $\dagger$ & \multicolumn{1}{c|}{2.7845 / 2.1741 / 4.7823} & \multicolumn{1}{c|}{35.66 / 13.85 / 14.05} & 0.98   & \textbf{9.6} & \textbf{0.076} & \textbf{0.25} \\

IOANet w/o attention & \multicolumn{1}{c|}{2.7731 / 2.3008 / \textbf{4.3190}} & \multicolumn{1}{c|}{35.35 / 13.82 / \textbf{14.12}} & 0.98   & \textbf{9.6} & \textbf{0.076} & \textbf{0.25} \\

IOANet & \multicolumn{1}{c|}{\textbf{2.3344} / \textbf{1.6414} / 4.6026 } & \multicolumn{1}{c|}{\textbf{36.84} / \textbf{13.86} / 14.11} & \textbf{0.99}  & 11.1 & \textbf{0.076} & \textbf{0.25} \\ \hline

IOANet $\ddagger$ & \multicolumn{1}{c|}{\textbf{1.7893} / \textbf{1.0937} / \textbf{4.0659} } & \multicolumn{1}{c|}{\textbf{38.76} / \textbf{14.08} / \textbf{14.26}} & \textbf{0.99}  & 11.1 & \textbf{0.076} & \textbf{0.25} \\ \hline

BEDSR \cite{lin2020bedsr} & \multicolumn{1}{c|}{--- / --- / ---} & \multicolumn{1}{c|}{--- / --- / ---} & ---  & OOM & OOM & $>>1K$ \\

LP-IOANet  & \multicolumn{1}{c|}{\textbf{1.8003 / 1.1259 / 4.0074}} & \multicolumn{1}{c|}{\textbf{38.69 / 14.08 / 14.33}} & \textbf{0.99} & \textbf{12.1} & \textbf{0.35} & \textbf{1.47} \\ 

\end{tabular}%
}
\vspace{-4mm}
\caption{Rows 2-5 (trained on BSDD) show the results of BEDSR, IOANet without attention and IOANet, all in \textit{low-resolution} ($192\times256$). Row 6 ($\ddagger$) shows IOANet trained on all three datasets (results in \textit{low-resolution}). The last two rows show BEDSR and LP-IOANet (trained on all three datasets), all in \textit{high-resolution} ($768\times1024$). $\dagger$ indicates training with only L1 loss. Runtime/memory are measured using an RTX 3090 GPU. Note that BEDSR runs out of memory (OOM) in high-resolution.}
\vspace{-4mm}
\label{tab:results}
\end{table*}

 \begin{table}[htbp!]
 \tiny
\resizebox{0.48\textwidth}{!}{%
\begin{tabular}{l|c|c}
Components & Resolution & Runtime (ms) \\  \hline

IOANet w/o attention &   $768\times1024$ &  80.3\\
IOANet & $768\times1024$ &  91.5 \\
LPTN \cite{liang2021high} + IOANet & $768\times1024$ & 142.6 \\
LPTN-lite + IOANet & $768\times1024$ & 84.7 \\
LP-IOANet & $768\times1024$ & \textbf{57.5 }\\
\end{tabular}%
}
    \vspace{-3mm}
    \caption{Runtime on a Samsung Galaxy S22 Ultra GPU. }
    \label{tab:mobile_timing}
\end{table}
\vspace{-5mm}

\begin{table}[!t]
\resizebox{0.48\textwidth}{!}{%
\begin{tabular}{l|c|c|cll}
\multicolumn{1}{c|}{\multirow{2}{*}{Upsampler}} & BSDD & \multicolumn{1}{c|}{Complexity} &  \multicolumn{2}{c}{Runtime (ms)} \\ \cline{2-5} 
\multicolumn{1}{c|}{} & MAE & GFLOPs & GPU & Mobile \\ \hline
LPTN \cite{liang2021high} & \textbf{1.7148} / \textbf{1.0594 }/ \textbf{3.8597} & 22.8 & 15.5&   142.6   \\
LPTN-lite & 1.7537 / 1.0784 / 3.964 & 3.82 & 14.2 &  84.70  \\
Ours (LP) & 1.8003 / 1.1259 / 4.0074 & \textbf{1.47} & \textbf{12.1} & \textbf{57.50}
\end{tabular}%
}
\vspace{-4mm}
 \caption{Different upsampling solutions with IOANet. LPTN \cite{liang2021high}, our LPTN-lite and LP-IOANet. Runtime values are taken on the same hardware as Tables \ref{tab:results} and \ref{tab:mobile_timing}.}
 \label{tab:upsampling}
 \vspace{-5mm}
\end{table}

\vspace{1.1mm}
\section{Experimental Results}
\label{experiments}
\vspace{-2mm}
\subsection{Implementation and Training Details} \label{sec:imp_details}
\noindent We adopt a two-stage training regime, where we first train the removal network IOANet (see Figure \ref{fig:networks}) in low-resolution and then train our upsampling framework LP-IOANet with the IOANet fixed (see Figure \ref{fig:overall_diagram}). We note that a one-stage, end-to-end training is possible, but since upsampling module requires high-resolution data, one-stage training can only be done on A-BSDD dataset. Two-stage training lets us train IOANet on low-resolution data and improves the final result.

In the first stage, we train IOANet for 1000 epochs using Adam \cite{kingma2015adam} with two losses; L1 and LPIPS \cite{zhang2018unreasonable}. Our loss weights are empirically chosen as 10 and 5 for L1 and LPIPS, and we use a mixed training strategy where we sample 15, 15 and 2 images in a batch from A-BSDD, Doc3DS+ and A-OSR, respectively. In the second stage, we freeze IOANet and train the upsampler using L1 loss. We train on A-BSDD for 200 extra epochs. Models are trained using PyTorch \cite{paszke2019pytorch}.

\vspace{-3mm}

\subsection{Evaluation Details and Results}
\noindent We use PSNR, SSIM and MAE metrics  on BSDD dataset for evaluation. We choose BSDD since we can train BEDSR on it (i.e. background colors are available) and because it is the only high-resolution dataset. We report metrics for all, non-shadow and shadow regions separately. We compare with the state-of-the-art BEDSR \cite{lin2020bedsr}; we reproduce the implementation and train it on A-BSDD.  We perform evaluation both in low-resolution ($192\times256$) and high-resolution ($768\times1024$).  

Our results are shown in Table \ref{tab:results}. We first compare IOANet, its components and BEDSR in low-resolution and train them only on BSDD. IOANet comfortably outperforms BEDSR on all metrics, despite running x10 faster, consuming x30 fewer memory and having x2K times fewer flops. It is also apparent that the input-output attention proposed in our architecture improves the results, with minimal to no overhead in runtime/memory performance. When trained on all three datasets (i.e. full stage 1 training), IOANet shows dramatic improvements, showing the value of our datasets. 

\begin{figure}[!t]
      \includegraphics[width=0.48\textwidth]{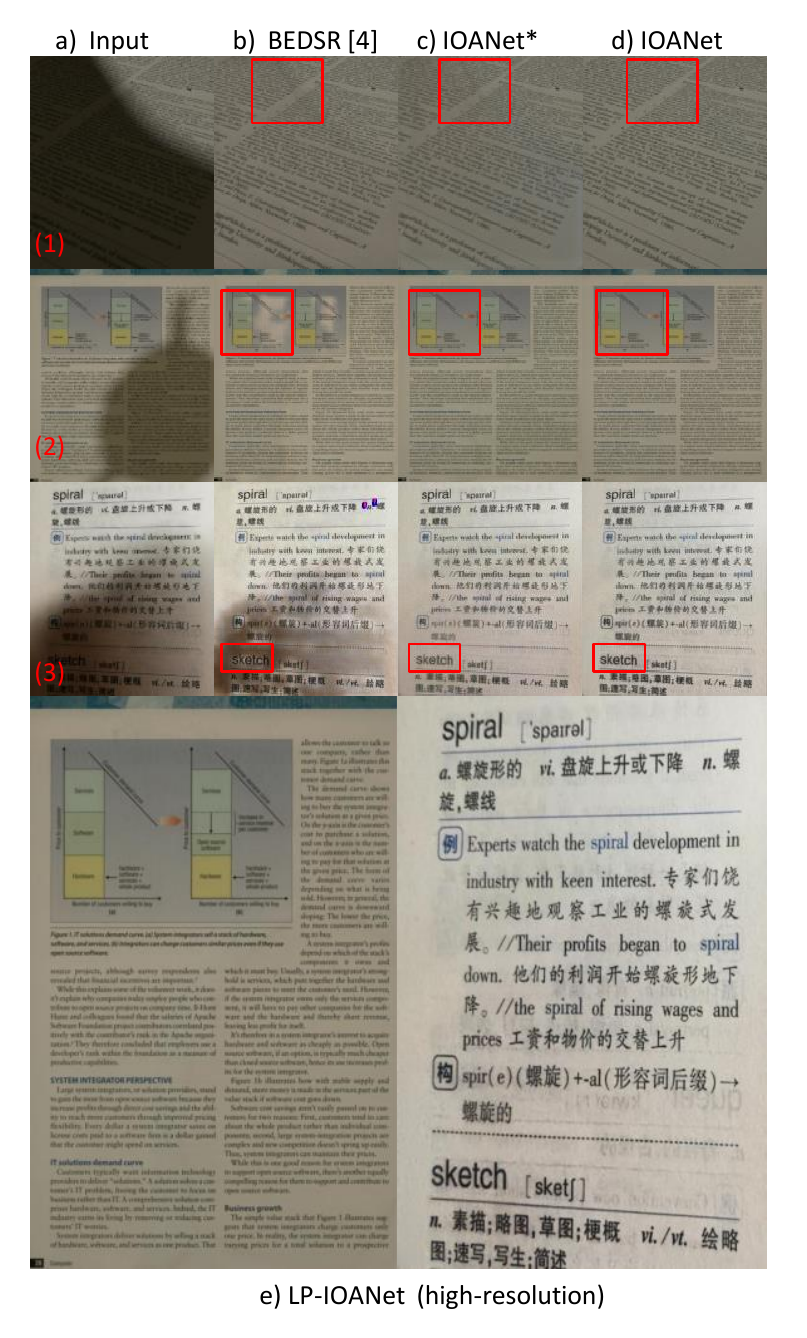}
      \vspace{-11mm}
      \caption{Visualization of a) input image, and output of b) BEDSR \cite{lin2020bedsr}, c) IOANet without attention (marked with *), d) our IOANet and e) our LP-IOANet. Differences are shown with red boxes, better viewed when zoomed in.}
      \vspace{-5mm}
  \label{qualitative}
\end{figure}

In high-resolution, we follow the full two-stage training and the results are even better; BEDSR fails to run in high-resolution and goes out-of-memory on a 24GB VRAM desktop GPU. Our LP-IOANet, on the other hand, achieves comparable results to our IOANet, despite evaluating at four times the resolution. Furthermore, it is still operating comfortably in real-time (around 84 FPS) while being still faster, smaller and less complex than low-resolution BEDSR. Figure \ref{qualitative} shows that our method handles artefacts (1st, 2nd row) and preserves high-frequency content (3rd row), even in high resolutions. Table \ref{tab:mobile_timing} shows that LP-IOANet reaches up to 20 FPS on mobile devices; it is faster than running IOANet directly on high-resolution, or using LPTN with IOANet.

\noindent \textbf{Further comparison.} Not many methods focus on runtime performance \cite{kligler2018document,lin2020bedsr,yang2012shadow,brown2006geometric}. Various non-ML methods \cite{bako2016removing, jung2018water,kligler2018document,osrDataset} have hardware-optimized implementations, but none of them utilize GPUs or can scale their results with more data. Our method leverages GPUs and scales its results significantly with more data (see Table \ref{tab:ablationDatasets}).

\vspace{-4mm}
\subsection{Ablation Studies}

\noindent \textbf{Upsampling.} Table \ref{tab:upsampling} shows three upsampling solutions with different complexities. There is an expected trend here; as the upsampler becomes more complex, we get better results. The timings on desktop GPUs are not that different, but the difference becomes more visible on mobile devices, as LP-IOANet is nearly three times faster than using the original LPTN, while being as accurate. We note that all variants perform in high-resolution and comfortably outperform BEDSR.

\noindent \textbf{Datasets.} \textcolor{black}{ Table \ref{tab:ablationDatasets} shows the contribution of each dataset on IOANet performance. Each dataset in the mix introduces visible improvements, despite naive mixing during training.}

\noindent \textbf{Loss terms.} The second (L1-only) and third rows (L1 + LPIPS) of Table \ref{tab:results} shows that using LPIPS as a loss term improves the results and justifies its addition.

 \begin{table}[htbp!]
\resizebox{0.48\textwidth}{!}{%
\begin{tabular}{l|c|c}
Training Dataset & MAE & PSNR \\ \hline
BSDD & 2.3344 / 1.6414 / 4.6026 & 36.84 / 13.76 / 14.11	\\
A-BSDD & 2.1163 / 1.4418 / 4.4220 & 37.43 / 13.91 / 14.14	 \\
BSDD \& Doc3DS+ & 2.0560 / 1.3761 / 4.2813 & 37.73 / 13.99 / 14.18	 \\
All & \textbf{1.7893 / 1.0937 / 4.0659}  & \textbf{38.76 / 14.08 / 14.26}	
\end{tabular}%
}
\vspace{-3mm}
    \caption{BSDD evaluation when IOANet is trained with different datasets. \textit{All} refers to A-BSDD, A-OSR and Doc3DS+.}
    \label{tab:ablationDatasets}
\end{table}

\vspace{-2mm}

\vspace{-4mm}
\section{Conclusion}
\label{conclusion}
\vspace{-2mm}

\textcolor{black}{We propose LP-IOANet, an end-to-end, lightweight high-resolution document shadow removal solution. It consists of IOANet, a shadow removal architecture, encapsulated within a lightweight upsampler. We also propose three new datasets, which helps generalize our model to in-the-wild scenarios. Our results show that LP-IOANet comfortably outperforms the existing state-of-the-art, while running on a mobile-device in real-time in high resolution.}

\clearpage

\bibliographystyle{IEEEbib}
\bibliography{strings,refs}
\end{document}